\pdfoutput=1
\documentclass[10pt, conference, compsocconf]{IEEEtran}
\usepackage{cite}

\usepackage[cmex10]{amsmath}
\usepackage{array}
\usepackage{mdwmath}
\usepackage{mdwtab}
\usepackage[caption=false,font=footnotesize]{subfig}
\usepackage{fixltx2e}
\usepackage{stfloats}
\usepackage{url}
\usepackage{color}
\usepackage{amssymb}
\usepackage{booktabs} %For the table
\usepackage{multirow} %For the table
\def\argmin{\mathop{\arg \min}}
\newcommand\redcomment[1]{\textcolor{black}{#1}}

\newcommand\finalcomment[1]{\textcolor{black}{#1}}

\usepackage{tikz}
\usepackage{lipsum}
\usepackage{textcomp}

\newcommand\copyrighttext{%
  \footnotesize To appear in proccedings of IEEE Pattern Recognition in Neuroimaging (PRNI) 2015. DOI 10.1109/PRNI.2015.10 
\textcopyright 2015 IEEE. Personal use of this material is permitted. Permission from IEEE must be obtained for all other uses, in any current or future media, including reprinting/republishing this material for advertising or promotional purposes, creating new collective works, for resale or redistribution to servers or lists, or reuse of any copyrighted component of this work in other works. }
\newcommand\copyrightnotice{%
\begin{tikzpicture}[remember picture,overlay]
\node[anchor=south,yshift=10pt] at (current page.south) {\fbox{\parbox{\dimexpr\textwidth-\fboxsep-\fboxrule\relax}{\copyrighttext}}};
\end{tikzpicture}%
}

\begin{document}
%
% paper title
% can use linebreaks \\ within to get better formatting as desired
\title{Kernel convolution model for decoding sounds from time-varying neural responses}

% author names and affiliations
% use a multiple column layout for up to two different
% affiliations
 \author{%\IEEEauthorblockN{Authors}
 \IEEEauthorblockN{\redcomment{Ali Faisal, Anni Nora, Jaeho Seol, Hanna Renvall and Riitta Salmelin}}
 \IEEEauthorblockA{Department of Neuroscience and Biomedical Engineering, \\
 Aalto University, Finland\\
 Email: ali.faisal@aalto.fi and riitta.salmelin@aalto.fi\\}}

\maketitle

\copyrightnotice
\begin{abstract}
In this study we present a kernel based convolution model to characterize neural responses to natural sounds by decoding 
their time-varying acoustic features. The model allows to decode natural sounds from high-dimensional neural recordings, such as 
magnetoencephalography (MEG), that track timing and location of human cortical signalling noninvasively across multiple channels. 
We used the MEG responses recorded from subjects listening to acoustically different environmental sounds. By decoding the stimulus 
frequencies from the responses, our model was able to accurately distinguish between two different sounds that it had never encountered 
before with 70\% accuracy. Convolution models typically decode frequencies that appear at a certain time point in the sound signal by using 
neural responses from that time point until a certain fixed duration of the response. Using our model, we evaluated several fixed durations 
(time-lags) of the neural responses and observed auditory MEG responses to be most sensitive to spectral content of the sounds at 
time-lags of 250~ms to 500~ms. The proposed model should be useful for determining what aspects of natural sounds are represented by 
high-dimensional neural responses and may reveal novel properties of neural signals.

\end{abstract}

% \begin{IEEEkeywords}
% neural decoding; kernel regression; MEG;
% 
% \end{IEEEkeywords}

% For peer review papers, you can put extra information on the cover
% page as needed:
% \ifCLASSOPTIONpeerreview
% \begin{center} \bfseries EDICS Category: 3-BBND \end{center}
% \fi
%
% For peerreview papers, this IEEEtran command inserts a page break and
% creates the second title. It will be ignored for other modes.
\IEEEpeerreviewmaketitle

\section{Introduction}
The way our brain represents periodic signals in different sensory modalities has been a subject of several studies. 
For example, spiking of movement-sensitive neurons in response 
to periodic signals was successfully encoded using the convolution model \cite{bialek1991} \redcomment{which} is a linear mapping 
from time-varying neural responses to time-varying 
representation of the incoming stimuli. The model has been subsequently employed \redcomment{in many studies} e.g. to 
investigate how 
the primary auditory cortex neurons encode 
spectro-temporal features in invasive recordings of ferrets \cite{mesgarani2009} and humans \cite{pasley2012}\redcomment{, to study the robustness %of the neural code \cite{mesgarani2014} 
and the extent 
to which perceptual aspects are coded in the cortical representation %\cite{mesgarani2014,
\cite{mesgarani2012}, and to characterizing 
stimulus-response function of auditory neurons \cite{calabrese2011}. }

Earlier studies addressing the spectro-temporal encoding in the human auditory system have typically used invasive intracortical 
recordings with limited spatial coverage.
For studying the $\mbox{spatio-temporal}$ response across the entire cortex one can utilize MEG which can track the timings and location 
of cortical responses at high resolution.
However, direct application of the convolution model to MEG \redcomment{data} is computationally challenging, as the 
complexity of the model is directly proportional to the spatial dimensionality of the neural response data, which is usually 
very high in MEG.
In this study we propose the {\em kernel convolution model}, which is a dual 
representation of a sparse convolution model and 
has an efficient parameter estimation scheme that is independent of the 
spatial dimensionality of neural responses.
We first show that the presented methodology using time-varying acoustic 
features of sound stimuli, here spectrogram, is able to 
decode new sounds with high accuracy. 
We then evaluate different $\mbox{time-lags}$ of the MEG responses in decoding 
the spectrogram of \redcomment{test} sounds in a cross-validation setting.
% 
% \subsection{Subsection Heading Here}
% Subsection text here.
% 
% 
% \subsubsection{Subsubsection Heading Here}
% Subsubsection text here.

\section{Convolution based predictive modelling}
%In this section we formalize the notion of the \redcomment{kernel} convolution model. 
\begin{figure}[!t]
\centering
\includegraphics[width=3.3in]{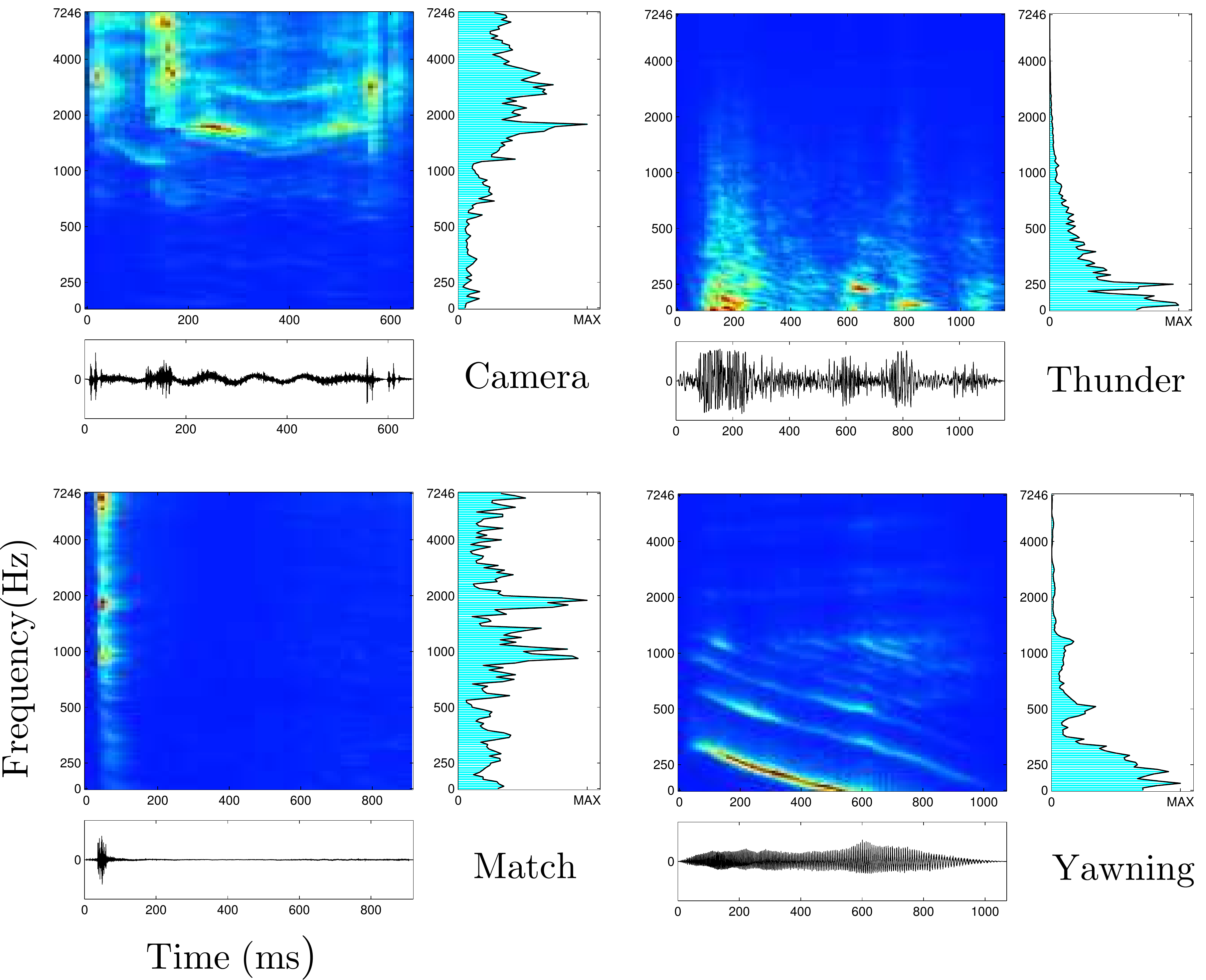}
\caption{Spectrograms and fourier transforms of four sample sounds.
The amplitude waveform of the sound is depicted below each spectrogram.}
\label{fig_spec}
\end{figure}
% %: camera, thunder, lighting a match and yawning. 
%, chosen from semantic categories vehicle, music, animal and human, respectively. %Spectrogram of each sound, with the sound waveform (below) and Fourier 
%transform (right).}
%(Left) spectrogram of the sound, (Botton) waveform of the sound and (Right) fourier transform of the sound.}% with frequencies matched to the central frequencies of the frequency bins in spectrogram.}

\subsection{Convolution model}
The convolution model \cite{bialek1991, mesgarani2009, pasley2012} is a linear mapping between the response of a population of 
neurons and a time-varying representation of the original stimulus, 
\redcomment{here spectrogram $s(t,f)$, sampled at 
times $t=1,...,T$ (see Fig.~\ref{fig_spec} for the spectrograms of four example sounds used in the study). The mapping is performed via a convolution of the 
neural responses evoked by the sound $r(t,x)$ with unknown $\mbox{spatio-temporal}$ response functions $g(\tau, f, x)$}
% In this model, the reconstruction weights (or the $\mbox{spatio-temporal}$ 
% response function) $g(\tau, f, x)$ are computed to map the neural population responses, $r(t,x)$, sampled at 
% times $t=1,...,T$ back to the sound spectrogram $s(t,f)$ under an assumed Gaussian noise model
\begin{equation}\label{eq:orig}
\hat{s}(t,f) = \sum_x\sum_{\tau} g(\tau, f, x) r(t-\tau, x) + \epsilon, 
\end{equation}
where $x$ indexes the MEG vertices (here sensors), $f$ represent the frequency channels, $\tau$ indicates the fixed 
duration \redcomment{(also referred to as the temporal lag above),} and $\epsilon$ is an 
additive zero~mean Gaussian random variable. 
\redcomment{In this model,} 
the reconstruction for each frequency channel $\hat{s}_f$ is treated independently of the other channels. 
If we consider the reconstruction of one channel, it can be written as
\begin{equation}\label{eq:orig_freq}
\hat{s}_f(t) = \sum_x\sum_{\tau} g_f(\tau, x) r(t-\tau, x) + \epsilon.
\end{equation}
To simplify the description of the inference algorithm used in this study, we transform the model in a linear algebraic form. First 
we define the response matrix \redcomment{$R \in \mathbb{R}^{N T \times \tau x}$}, such that each row $r_n(t)$ contains the MEG response profile to sound $n$ across the entire 
set of sensors $x$ at time $t$ and the subsequent $\tau$ time bins:
\begin{align}
&R = \nonumber \\
&\begin{bmatrix}
	r_1(1,1) & r_1(1,2) & \cdots & r_1(1,x) & \cdots & r_1(1-\tau,x)\\
	r_1(2,1) & r_1(2,2) & \cdots & r_1(2,x) & \cdots & r_1(2-\tau,x)\\
	\vdots  & \vdots  & \ddots & \vdots \\
	r_1(T,1) & r_1(T,2) & \cdots & r_1(T,x) & \cdots & r_1(T-\tau,x)\\
	\vdots  & \vdots  & \vdots \\
	r_N(T,1) & r_N(T,2) & \cdots & r_N(T,x) & \cdots & r_N(T-\tau,x)\\
\end{bmatrix}\nonumber
\end{align}
  %\left[
  %\right]
\begin{equation}
G_f = \begin{bmatrix} g_f(1,1) & g_f(1,2) & \cdots & g_f(1,x) & \cdots & g_f(\tau,x) \end{bmatrix}^\top\nonumber\\
\end{equation}
and
\begin{equation}
 S_f = \begin{bmatrix}
        s_{f}(1,1) & s_{f}(1,2) & \cdots & s_{f}(1,T) & \cdots & s_{f}(N,T)
       \end{bmatrix}^\top\nonumber
\end{equation}
Using the matrix notation, Eq~\ref{eq:orig_freq} becomes: $S_f = R G_f + \epsilon$, 
which is similar to multiple linear regression with weights $G_f$.
Given a pre-defined lag, %\footnote{The value of the lag $\tau$ ranges from $-T$ to $0$ and both $T$ and $\tau$ 
%are of the same order.}, 
the function $G_f$ is estimated by minimizing 
the mean-squared error between the actual and the predicted stimuli: 
%\begin{equation}
$\argmin_{G_f}\sum_{n,t}\{s_f(n,t)-\hat{s}_f(n,t)\}^2$. 
%\end{equation}
Solving this results in a maximum likelihood (ML) estimate:
\begin{equation}\label{ml}
\hat{G_f} = (R^\top R)^{-1} R^\top S_f.
\end{equation}
The estimate requires an inversion of the inner product $R^\top R$ \redcomment{that} has a dimension 
$d \times d$, where $d = \tau x$ is the dimension of the MEG response data. In neuroimaging, 
particularly in MEG, the value of $d$ is typically large. This is primarily due to the high spatial resolution of 
MEG where the data is sampled from hundreds to thousands \redcomment{vertices}, $x$, depending on whether the data is 
represented at the sensor- or source-level. 
Further, the different sources can be highly correlated in MEG which makes the inversion $\mbox{ill-conditioned}$, i.e., the 
resulting inverse may not be possible to compute or it may be very sensitive to slight variation in the data.

\begin{figure}[!t]
\centering
\includegraphics[width=3.0in]{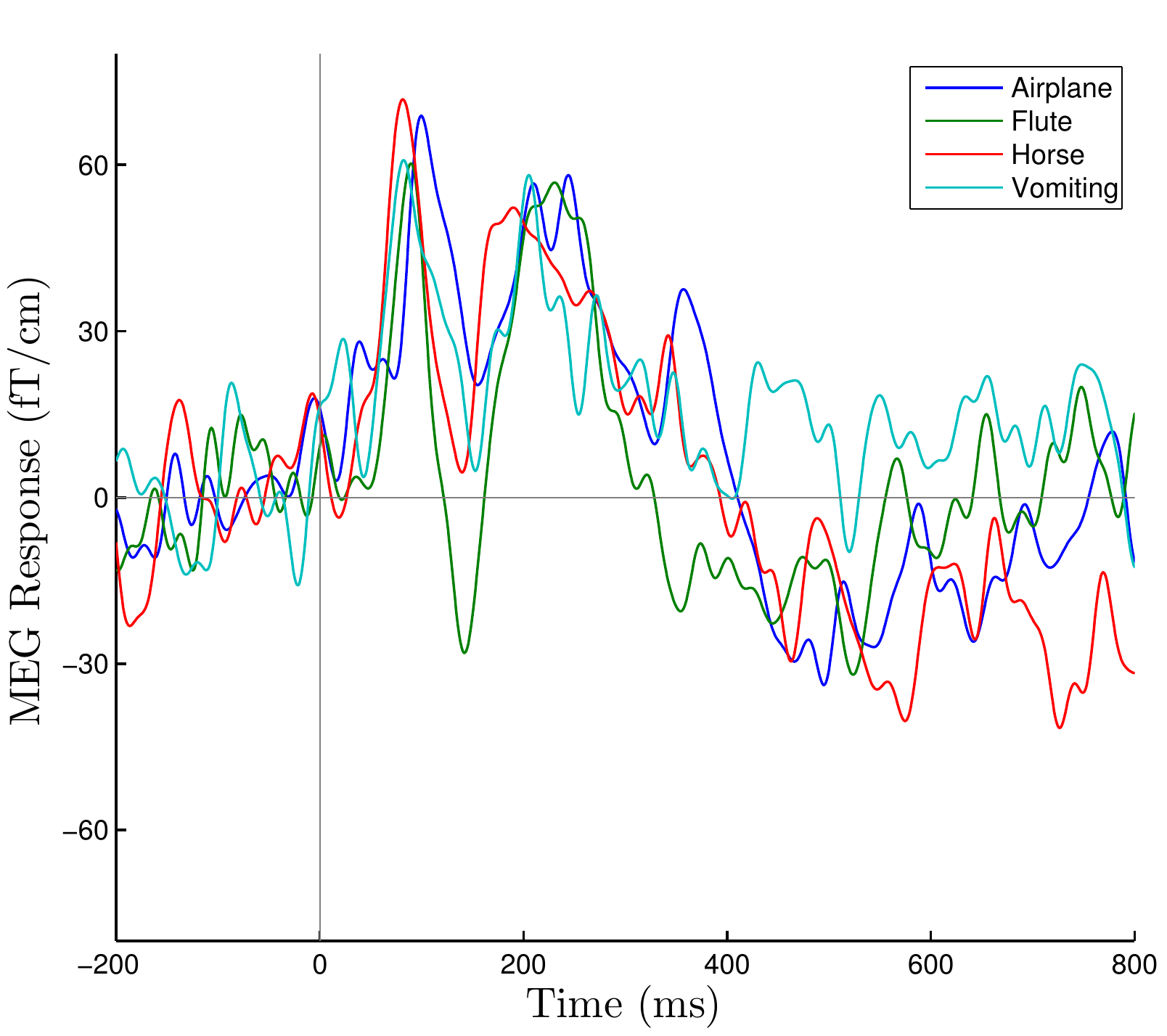}
\caption{Event-related responses in one subject (averaged over 20 repetitions of the sound) at an MEG 
sensor located over the left hemisphere.}
\label{fig_sample_ER}
\end{figure}
 %from a left temporal channel recorded from the first subject.}

\subsection{Kernel convolution model}
By \redcomment{applying} similar developments for linear regression \cite{hastie2001}, we %,shawe2004} we 
reformulate the classical convolution model in terms of its kernel or dual representation and 
add suitable regularization. 
%The kernel representation avoids inverting the inner product. 
In this representation, we use a sparse prior on the response function $G_f: G_f\sim N(0,\lambda_f^{-1}\mathrm{I})$, 
where $\lambda_f \ge 0$ is the regularization parameter and $\mathrm{I}$ is an identity matrix. 
%with diagonal entries set to one and all other entries set to zero. 
The function can be determined by maximizing the $\mbox{log-posterior}$ distribution of $G_f$ which is equivalent to 
minimizing the regularized sum-of-squares error function given by 
%representation of the convolution model which can be seen 
%as linear regression with a penalty term penalizing high values for the function $g$. The function can be determined by 
%minimizing the regularized sum-of-squares error function given by
\begin{equation}\label{map_kernel}
\argmin_{G_f} \sum_{n,t}\{s_f(n,t)-\hat{s}_f(n,t)\}^2 + \lambda_f\sum_{x,\tau} g_f(\tau,x)^2.
\end{equation}
%The minimum can be found by setting the gradient with respect to $G_f$ to zero which 
Solving this yields a maximum a posteriori (MAP) estimate:
\begin{align}\label{map}
\hat{G_f} =&  (R^\top R + \lambda_f \mathrm{I})^{-1} R^\top S_f.
\end{align}
The addition of the 
regularization term stabilizes the estimation of the inverse. Following the derivation of 
kernel ridge regression \cite{bishop2006}, the MAP estimate can be obtained using the dual form of the sparse convolution model:
\begin{align}\label{kernel}
\hat{G_f}=&  R^\top (R R^\top + \lambda_f \mathrm{I})^{-1} S_f.
\end{align}
Unlike the original form \redcomment{(Eq.~\ref{map} or 
the non-sparse version: Eq.~\ref{ml})} that required the inversion of $R^\top R \in \mathbb{R}^{(\tau x) \times (\tau x)}$, the dual form requires inversion of the Gram matrix 
$K=R R^\top \in \mathbb{R}^{(N T) \times (N T)}$. This is very useful for neuroimaging 
studies where the number of conditions, $N$, is typically very low compared to the number of neural 
sources $x$ while $\tau$ and $T$ are of the same order.
%Both the ML estimate (in Eq.~\ref{ml}) and the MAP estimate (in Eq.~\ref{map}) are the well known solutions of linear 
%regression problem \cite{hastie2001, bishop2006}. 
To estimate $\lambda_f$, we follow \cite{sudre2012} and use an efficient computational technique \cite{guyon2005}, which avoids the inversion $(R R^\top + \lambda_f \mathrm{I})^{-1}$ for each 
value of $\lambda_f$ and uses a fast scoring measure to estimate leave-one-out error for different values of the 
regularization parameter.

%The sparse kernel model allows fast estimation of the convolution functions $G_f$. 
The entire reconstruction of the \redcomment{sound} 
spectrogram can be described as the collection of convolution functions for each frequency channel; 
$\hat{G}=[\hat{G_1} \hat{G_2} \cdots \hat{G_F}]$. Then, given a\redcomment{n} MEG response to a test sound, we take the lagged representation \redcomment{of the response}, 
$r_{\text{new}} \in \mathbb{R}^{T\times (\tau x)}$, and obtain a prediction of its spectrogram $\hat{S}_{\text{new}}\in \mathbb{R}^{T\times F}$ as follows:
\begin{align}\label{ker1}
\hat{S}_{\text{new}} = r_{\text{new}}\hat{G} = r_{\text{new}} R^\top (R R^\top + \lambda_f \mathrm{I})^{-1} S.
%&= r_{\text{new}} R^\top (R R^\top + \lambda_f \mathrm{I})^{-1} S.
\end{align}
\redcomment{The dual formulation can be obtained by noticing that} the 
prediction in Eq.~\ref{ker1} operates on the feature space and only involves inner products. These inner~products can be replaced with a kernel 
function $k(r_n, r_m)=\mathbf{\phi}(r_n)^\top \mathbf{\phi}(r_m)=\sum_i\phi_i(r_n)\phi_i(r_m)$, where 
$\phi_i(r)$ are the basis functions. If we substitute the kernel functions for the inner-products we obtain the 
following prediction of the spectrogram:
%\begin{equation}
$\hat{S}_{\text{new}} =\text{k}(r_{\text{new}})(K + \lambda_f\mathrm{I})^{-1} S$, 
%\end{equation}
where we have defined the matrix $\text{k}(r_{\text{new}})$ with column-wise concatenation of submatrices $k(r_{\text{new}}, r_{n})$. 
Similarly, the submatrices of $K$ are defined using the kernel function $k(r_n,r_m)$.
Thus, the dual formulation implicitly allows to use feature spaces of very high, even infinite, dimensions.

\subsection{Model evaluation}
We performed a leave-two-out cross-validation where, in each fold, we used all but two randomly picked sounds as
training data. 
\redcomment{To label the held-out sounds without using any training examples for those sounds, we followed a two-stage 
prediction procedure, similar to} \cite{mitchell2008}. 
\redcomment{In the first stage,} we 
applied the learned functions to predict the spectrograms 
for the test sound \redcomment{pair and concatenated the temporal dimension to form vectors for both 
predicted and original spectrograms.} \redcomment{In the second stage,} we quantified 
the predictive accuracy by computing the correlation between the reconstructed and 
the original spectrogram of the two test sounds. If the two predictions are represented as 
$p_1$ and $p_2$ and the original spectrograms are 
$s_1$ and $s_2$, then the labelling assigned by the model was considered correct if: 
\begin{equation}\label{pairwise-dist}
\mbox{corr}(s_1, p_1) + \mbox{corr}(s_2, p_2) > \mbox{corr}(s_1, p_2) + \mbox{corr}(s_2, p_1) 
\end{equation}
This process is repeated for all possible combinations of leave-two-out sounds.
% and is similar to the evaluation schemes followed for decoding words and pictures \cite{sudre2012}.
Under this evaluation, the expected performance of a random model is $50\%$. \redcomment{Since the sounds 
are of different durations, to evaluate Eq.~\ref{pairwise-dist}, we truncated the 
predicted and original spectrogram to the length of the shorter sound in each test sound pair.} 
%Note that this truncation can only underestimate the model's performance.

To evaluate how well the spectrogram features were predicted, we use the following score:%\redcomment{coefficient of determination score \cite{steel1960}} 
\begin{equation}\label{score}
\mbox{score}_{f,t} = 1 - \frac{\sum(s_{f,t} - \hat{s}_{f,t})^2}{\sum(s_{f,t} - \bar{s}_{f,t})^2},
\end{equation}
where $s_{f,t}$ is the original value of the spectrogram frequency $f$ at time $t$, $\hat{s}_{f,t}$ is the predicted value by the model, 
and $\bar{s}_{f,t}$ is the mean value across all pairs in the cross-validation combinations. The summations in Eq.~\ref{score} 
are computed over all pairs of cross-validation samples. The feature score thus measures percent of variation 
explained in each feature. %A value close to one indicates a \redcomment{good} prediction of the feature. 
%\begin{compactitem}
%\item Donec dolor
%\end{compactitem}

\begin{figure}[!t]
\centering
\includegraphics[width=3.3in]{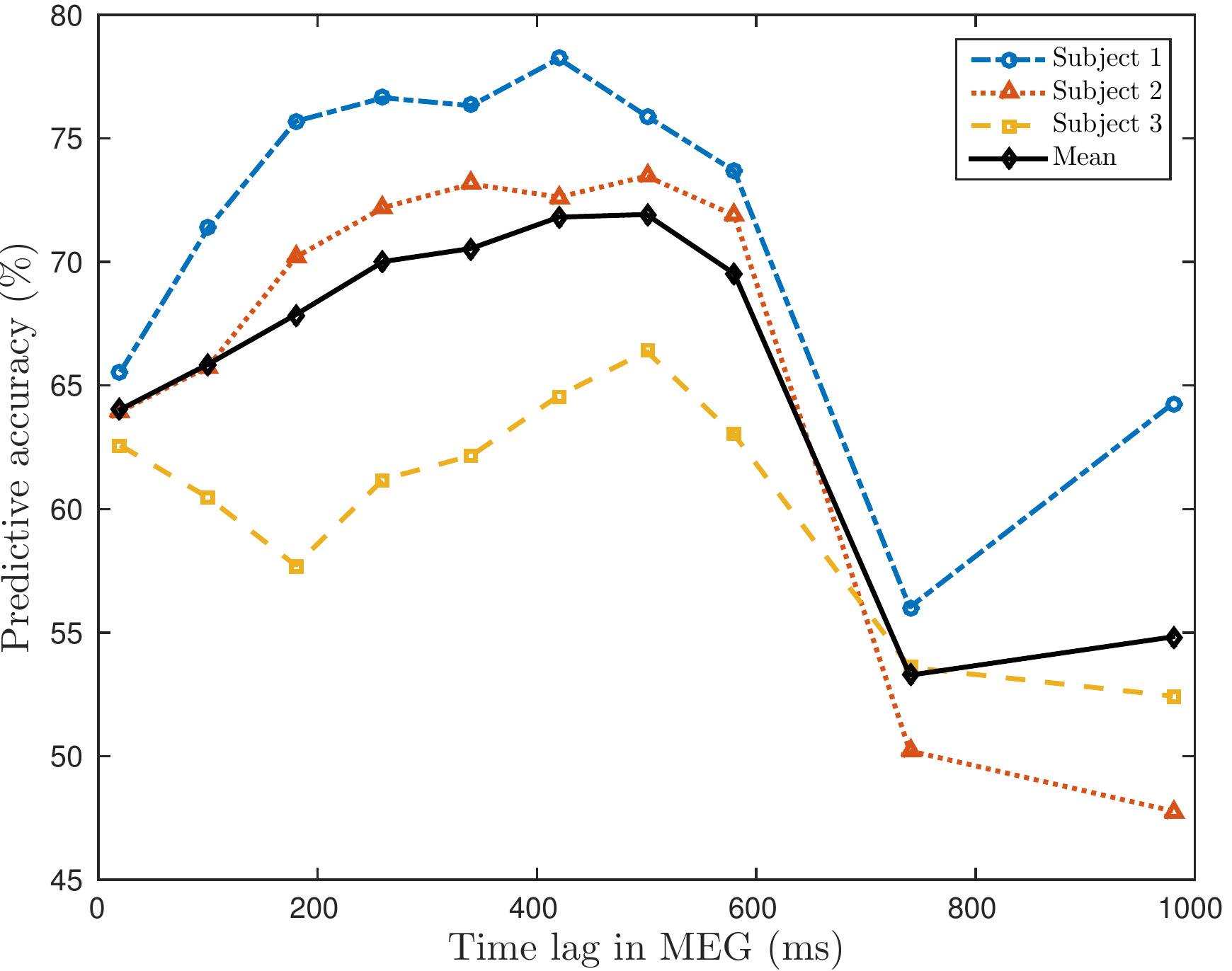}
\caption{Performance of kernel convolution model as a function of time-lag. Each point is an average over 946 
cross-validation tests. Chance accuracy is 50\%. Solid line is average accuracy across the three subjects.}
\label{fig_perf}
\end{figure}

%------------------------------------------------
\section{Experiments}
\subsection{MEG recordings}
The data consisted of event-related MEG responses from three subjects listening to common environmental sounds (\redcomment{$44$} items). 
The sounds included sets of $6 - 8$ items from five pre-selected categories (vehicles, music, human, animal and 
tool) and $8$ uncategorized sounds. %and 8 nonmeaningful sounds. 
Each sound was presented $20$ times. %\redcomment{The protocol for MEG recordings was approved by Aalto University ethical committee.}
%The sounds \redcomment{$44$} 
%were tested in a independent behavioral tests which validated that users can detect the item by listening to its sound.
% Additional foil 
% sounds were presented $1 - 3$ times during measurement, and not used for further analysis. After the measurements, the participants 
% performed an independent behavioral task to check whether they recognized the categories of each $52$ sound during the MEG measurement. 

Magnetic fields associated with neural current flow were recorded with a $306$-channel whole-head neuromagnetometer 
(Elekta Oy, Helsinki) in the Aalto NeuroImaging MEG Core. 
The MEG signals were band-pass filtered between $0.03$ and $330$~Hz and sampled at $1000$~Hz. 
%The measurements were performed in two 
%$40$~min sessions for each participant, paused at $5$ minute intervals.
During the recordings subjects listened to a pseudo-randomly 
shuffled sequence of sounds and were %engaged in a one-back task where they were 
asked to respond by finger lift when two 
consecutive sounds referred to the same item. Response trials were excluded from analysis. 
% Each item consisted of $20$ repetitions during measurement of two sessions, and each 
% session takes about 40 minutes with pauses at $5$ minute intervals. The entire measurement for a single session lasted for 
% 1.5 hours. Item-level MEG responses were collected between $0$ and $2$ seconds from stimulus onset. These responses were 
% down-sampled to 10 ms-interval and were used to compute sensor-level evoked responses across the $20$ repetitions. The 
% $52$ sounds include a set of 6-8 items each collected from five pre-selected semantic categories (vehicles, music, human, 
% animal and tool), 8 filler sounds (not categorized) and 8 non-meaningful sounds. We also included additional foil sounds 
% presented 1 - 3 times during measurement, and not used for the further analysis. After the measurements, the participants 
% performed an independent behavioral task to check whether they recognized the categories of each 52 sound during the MEG 
% measurement.
The event-related responses to the $20$ repetitions of each stimulus were averaged from $300$~ms before to $2000$~ms after the 
stimulus onset, rejecting trials contaminated by %artifacts (signal strength in planar gradiometers exceeding $3000$~fT/cm). 
eye movements. 
%Analysis of the data was restricted to planar gradiometer channels above the bilateral auditory cortices, resulting in $56$~channels. 
On average $19.2 \pm1.1$ (mean $\pm$ standard deviation) artifact-free epochs \redcomment{(repetitions)} per subject were gathered for each item. 
%(maximum = $20$ repetitions).
The averaged MEG responses were baseline-corrected to the $200$~ms interval immediately preceding the stimulus onset and 
down-sampled to $10$~ms intervals. 
Data analysis was restricted to \redcomment{$56$} planar gradiometers
above the auditory cortex. % \redcomment{cortex in each hemisphere (total 56 sensors)}. 
Example responses are depicted in Fig.~\ref{fig_sample_ER}.

\subsection{Stimulus spectrogram representation}
% The auditory spectrogram is calculated based on the auditory filter bank with $128$ overlapping bandpass filter channels, with their 
% central frequencies ranging from $180$ to $7246$~Hz using the NSL toolbox (available at \url{http://www.isr.umd.edu/Labs/NSL/Software.htm}). 
The auditory spectrogram representation was binned at $10$~ms and calculated based on the auditory filter bank with $128$ overlapping 
bandpass filter channels 
mimicking the auditory periphery \cite{Chi2005}. Filters had logarithmically spaced central frequencies ranging from $180$ to $7246$~Hz (Fig.~\ref{fig_spec}). 
%The spectrograms were computed using the NSL toolbox (available at \url{http://isr.umd.edu/Labs/NSL/Software.htm}).
%Fig.~\ref{fig_spec} 
%shows the computed spectrograms and waveforms for four example sounds.

\subsection{Prediction of sound spectrograms from MEG responses}
\redcomment{Prior to the analysis, both spectrogram and MEG data were standardized to zero mean and unit variance. We used causal response functions ($\tau \le 0$; \cite{mesgarani2012}), which means that the model decoded spectrograms of sounds at time $t$
using neural responses at time $t, t+1, t+2, ..., t-\tau$~ms. } 
To evaluate the sensitivity of MEG neural responses to the frequencies in the stimulus spectrogram, 
%we performed the entire evaluation across % \redcomment{range of} values for the $\mbox{time-lags}$ \redcomment{($\tau$~=~$-20$~to~$0$, %$-100$~to~$0$,~$\cdots$, $-980$~to~$0$~ms)}.
we evaluated the mean 
predictive accuracy across all possible leave-two-out combinations of $44$ sounds ($C_{2}^{44} = 946$ combinations) for different 
time-lags $-\tau$~=~$20$, $100$, $180$, $260$, $340$, $420$, $500$, $580$, $740$ and $980$ ms.  
%This implies that for the $500$~ms lag ($\tau=-500$~ms), spectrograms of sounds at time $t$ are predicted using MEG responses at time: 
%$[t, t+1, t+2, ..., t+500]$~ms.
%For each time-lag, the mean 
%predictive accuracy across all possible leave-two-out combinations of $44$ sounds ($C_{2}^{44} = 946$ combinations) is 
%%The average predictive performance across all $C_{2}^{44} = 946$ possible leave-two-out combinations of meaningful sounds is 
Results, shown in Fig.~\ref{fig_perf}, indicate that it was possible to discriminate between two \redcomment{previously unencountered test} sounds with $\sim 70\%$ 
accuracy (Mean value $70.0$ to $71.9$ at $\mbox{time-lag}$ $250$ to $500$~ms) %based on their spectrograms predicted from observed 
%MEG responses 
even when neither sound \redcomment{was} used in the training data. 
%The figure shows that the neural responses are particularly sensitive to the frequencies from $250$~ms to about $500$~ms after 
%stimulus onset. 
%The spectrograms were particularly well predicted when time lags of $\sim250 - 500$ ms were used for the MEG data.
%\subsection{Predictions of sound spectrograms}
Next, to evaluate which spectrogram features were \redcomment{best} predicted, we considered 
the $\mbox{time-lag}$ of $500$~ms that gave the optimal predictions and computed 
%list of spectrogram features that were \redcomment{best} predicted by the model \finalcomment{across the three subjects.} 
\finalcomment{the mean score (Eq \ref{score}) across the three subjects for each spectrogram feature.} 
%Table~\ref{tab_scores} lists the top $15$ central 
%frequencies of the sounds with corresponding time in the spectrogram that \redcomment{had} 
%the highest mean feature score across 
%the three subjects. 
%These 
\finalcomment{The top $15$ scoring} features represent high stimulus frequencies (above $3.8$ kHz) \finalcomment{with scores 
ranging from $0.12$ to $0.21$}. 
\redcomment{Further, we computed $\mbox{item-wise}$ mean 
predictive accuracy over the cross-validation folds. The five best predicted sounds were camera $(95.3\%)$, helicopter $(88.4\%)$, lighting a match $(84.5\%)$, motorsaw $(83.7\%)$, and door $(82.9\%)$, while five least accurately predicted sounds were trumpet $(59.7\%)$, laughter $(58.9\%)$, yawning $(56.6\%)$, zipper $(55.0\%)$, and thunder $(54.3\%)$. The best predicted 
sounds typically contained higher frequencies compared to the less accurately predicted sounds (see Fig.~\ref{fig_spec}).}

% \begin{table}
% \caption{Top $15$ spectrogram features sorted by mean feature score calculated using data from three subjects. The spectrogram features are shown 
% as the central frequency of the spectral bin at a time from the start of the sound.}\label{tab_scores}%1892 all three subjects}
% \centering
% \begin{tabular}{lcc}
% \toprule
% \multirow{2}{*}{Mean score ($\pm$ S.D.)}& \multicolumn{2}{c}{Spectrogram feature} \\
% \cmidrule(r){2-3}
% & Frequency (kHz) & Time (ms) \\
% \midrule
% $ 0.21 $ $ (\pm 0.02 ) $ & $ 6.74 $&$ 30 $ \\
% $ 0.20 $ $ (\pm 0.02 ) $ & $ 6.94 $&$ 30 $ \\
% $ 0.18 $ $ (\pm 0.01 ) $ & $ 6.55 $&$ 30 $ \\
% $ 0.18 $ $ (\pm 0.09 ) $ & $ 6.36 $&$ 30 $ \\
% $ 0.16 $ $ (\pm 0.05 ) $ & $ 6.55 $&$ 70 $ \\
% $ 0.16 $ $ (\pm 0.14 ) $ & $ 3.89 $&$ 40 $ \\
% $ 0.15 $ $ (\pm 0.10 ) $ & $ 4.01 $&$ 40 $ \\
% $ 0.15 $ $ (\pm 0.05 ) $ & $ 6.36 $&$ 40 $ \\
% $ 0.15 $ $ (\pm 0.02 ) $ & $ 6.55 $&$ 60 $ \\
% $ 0.14 $ $ (\pm 0.01 ) $ & $ 6.55 $&$ 40 $ \\
% $ 0.13 $ $ (\pm 0.07 ) $ & $ 4.91 $&$ 50 $ \\
% $ 0.13 $ $ (\pm 0.05 ) $ & $ 6.74 $&$ 70 $ \\
% $ 0.13 $ $ (\pm 0.19 ) $ & $ 3.89 $&$ 50 $ \\
% $ 0.13 $ $ (\pm 0.01 ) $ & $ 6.74 $&$ 60 $ \\
% $ 0.12 $ $ (\pm 0.01 ) $ & $ 6.74 $&$ 40 $ \\
% \bottomrule
% \end{tabular}
% \end{table}
%------------------------------------------------

%\section{Results}

\section{Discussion and Conclusion}
Our results demonstrate that the kernel convolution model provides 
an efficient method for predicting spectrograms of new sounds. Predictions 
\redcomment{are made} by decoding neural information in high-dimensional MEG 
responses to common environmental sounds. 
\finalcomment{Therefore, the extracted neural information can be regarded 
as being based on neural mechanisms that generalize across a variety of sounds}. 
We evaluated different \redcomment{time-lags in} the MEG \redcomment{response} data to predict spectrograms of unencountered sounds, 
and observed that the responses are most sensitive \redcomment{for a duration of} around $250-500$ ms from the input stimulus. 
%\redcomment{The first cortical responses to auditory stimuli are not shown until approx. $15$~ms from the time the sound enters 
The auditory evoked responses used in the analysis are most prominent at $50-500$~ms after the stimulus 
onset despite the stimulus duration (see Fig.~\ref{fig_sample_ER}). Thus, %at the shortest calculated time lags 
%(here $20$~ms) the MEG responses are not informative enough for separating responses from each other, and 
at the longest $\mbox{time-lags}$ ($>500$~ms) \redcomment{the MEG data is noisier compared to shorter lags}, as the 
decaying MEG responses start to show large inter-response variability. %Additionally, the gradual decrease in performance 
%after $500$~ms may be due to the reduced coding duration ($< 500$~ms) of the neural response to 
\finalcomment{Neurophysiological interpretation and evaluation of significance of the results are natural 
extensions of the study. The decoding problem studied here is an example of an underdetermined systems for which 
regularization and Bayesian inference have 
provided reasonable answers.}%nonstationary sounds. 

Classical linear regression has been used earlier to decode neural responses, but most studies have either been limited to 
non-time-varying stimulus representations \cite{sudre2012} or neuroimaging recordings \cite{mitchell2008} \cite{santoro2014}. 
The proposed method will be useful for analyzing brain's ability to 
understand sounds in an acoustic environment, particularly when neural responses are recorded 
at high $\mbox{spatio-temporal}$ resolution.

\section*{Acknowledgment}
% We acknowledge the computational resources provided by the Aalto Science-IT project. % and the Finnish national IT Center for Science (CSC). 
% The work was supported by the Academy of Finland (LASTU programme $2012-2016$, personal grants to HR and RS) and the 
% Doctoral Program Brain and Mind (AN). The authors would like to thank Prof. Elia Formisano, Dr. Giancarlo Valente, Prof. Tom Mitchell 
% and Dr. Gus Sudre for valuable discussions.
We acknowledge the computational resources provided by the Aalto Science-IT project. % and the Finnish national IT Center for Science (CSC). 
\finalcomment{The work was supported by the Academy of Finland and the 
Doctoral Program Brain and Mind. We thank Elia Formisano, Tom Mitchell, Giancarlo Valente,  
and Gustavo Sudre for valuable discussions.}

% trigger a \newpage just before the given reference
% number - used to balance the columns on the last page
% adjust value as needed - may need to be readjusted if
% the document is modified later
%\IEEEtriggeratref{8}
% The "triggered" command can be changed if desired:
%\IEEEtriggercmd{\enlargethispage{-5in}}

% references section

% can use a bibliography generated by BibTeX as a .bbl file
% BibTeX documentation can be easily obtained at:
% http://www.ctan.org/tex-archive/biblio/bibtex/contrib/doc/
% The IEEEtran BibTeX style support page is at:
% http://www.michaelshell.org/tex/ieeetran/bibtex/
%\bibliographystyle{IEEEtran}
\bibliographystyle{IEEEtran}
\bibliography{main_cameraReady}
% argument is your BibTeX string definitions and bibliography database(s)
%\bibliography{IEEEabrv,../bib/paper}
%
% <OR> manually copy in the resultant .bbl file
% set second argument of \begin to the number of references
% (used to reserve space for the reference number labels box)
% \begin{thebibliography}{1}
% 
% \bibitem{IEEEhowto:kopka}
% H.~Kopka and P.~W. Daly, \emph{A Guide to \LaTeX}, 3rd~ed.\hskip 1em plus
%   0.5em minus 0.4em\relax Harlow, England: Addison-Wesley, 1999.
% 
% \end{thebibliography}

% that's all folks
\end{document}